\documentclass[letterpaper, 10 pt, conference]{ieeeconf}
\IEEEoverridecommandlockouts

\usepackage{graphicx}

\usepackage{amsmath,amsfonts,amssymb,amsthm}
\usepackage{array}

\usepackage[caption=false,font=normalsize,labelfont=sf,textfont=sf]{subfig}

\usepackage{url}

\usepackage[hidelinks]{hyperref}
\usepackage[ruled,linesnumbered]{algorithm2e}

\SetKwComment{Comment}{}{} 

\SetCommentSty{mycommfont}

\usepackage[noadjust]{cite}

\DeclareMathOperator*{\argmin}{\arg\!\min}
\begin{document}
\bstctlcite{IEEEexample:BSTControl}

\newcommand*{\vertbar}{\rule[-1ex]{0.5pt}{2.5ex}}
\newcommand*{\horzbar}{\rule[.5ex]{2.5ex}{0.5pt}}
\newcommand{\di}[1]{\mathrm{d}#1}

\title{Physics-informed Split Koopman Operators for Data-efficient\\Soft Robotic Simulation}

\author{Eron Ristich$^{1,2}$, Lei Zhang$^1$, Yi Ren$^1$, and Jiefeng Sun$^{1}$%
\thanks{$^1$School for Engineering of Matter, Transport and Energy at Arizona State University, Tempe, Arizona 85281, USA.  Email: \href{mailto:jiefeng.sun@asu.edu}{jiefeng.sun@asu.edu}}
\thanks{$^2$School of Computing and Augmented Intelligence at Arizona State University, Tempe, Arizona 85281, USA.}
}


\maketitle

\begin{abstract}


Koopman operator theory provides a powerful data-driven technique for modeling nonlinear dynamical systems in a linear framework, in comparison to computationally expensive and highly nonlinear physics-based simulations. However, Koopman operator-based models for soft robots are very high dimensional and require considerable amounts of data to properly resolve. 
Inspired by physics-informed techniques from machine learning, we present a novel physics-informed Koopman operator identification method that improves simulation accuracy for small dataset sizes. Through Strang splitting, the method takes advantage of both continuous and discrete Koopman operator approximation to obtain information both from trajectory and phase space data. The method is validated on a tendon-driven soft robotic arm, showing orders of magnitude improvement over standard methods in terms of the shape error. We envision this method can significantly reduce the data requirement of Koopman operators for systems with partially known physical models, and thus reduce the cost of obtaining data. More info: \url{https://sunrobotics.lab.asu.edu/blog/2024/ristich-icra-2025/}
\end{abstract}


%
\IEEEpeerreviewmaketitle

\section{Introduction}

Soft robots, owing to their continuous structure and compliant materials, can produce motions that closely mimic biological creatures. For example, a soft robotic arm can perform dexterous tasks in complex environments and operate more safely alongside humans than their rigid counterparts \cite{rus_design_2015}. As a result, soft robotic arms have been employed in a wide range of applications, including surgery \cite{cianchetti_soft_2014}, targeted drug delivery \cite{hu_small-scale_2018}, industrial tasks \cite{brown_universal_2010, wang_circular_2021}, and deep sea exploration \cite{li_self-powered_2021}. Despite of the advantages, it is challenging to model soft robotic arms due to the infinite degrees of freedom (DoFs) in their motions. In particular, the major difficulty is in predicting a soft robotic arm's dynamics with both high fidelity and high computation efficiency.


Most accurate physics-based models for soft robotic arms have high computational costs or nonlinearities that make it difficult to design effective controllers.
For example, general models based on continuum mechanics models such as Kirchhoff rod models \cite{bao_kinematics_2019} and Cosserat rod models \cite{rucker_statics_2011, till_real-time_2019, renda_dynamic_2014} inherently represent the continuous configuration space of the robot, but result in a system of partial differential equations (PDEs) that is computationally expensive to solve. Industrial finite element methods have also been employed in the case of control of the end-effector \cite{duriez_control_2013}. However, these methods often require high spatial and temporal resolution to capture the complex dynamics of soft robots, and as such are often too computationally expensive for real-time control. Reduced order models are a popular alternative that greatly improves computational speed of forward simulation \cite{goury_fast_2018, tonkens_soft_2021} while preserving accuracy. However, nonlinearities in resulting dynamics pose difficulties for controller design. 

\begin{figure}
    \centering    \includegraphics[width=.85\linewidth]{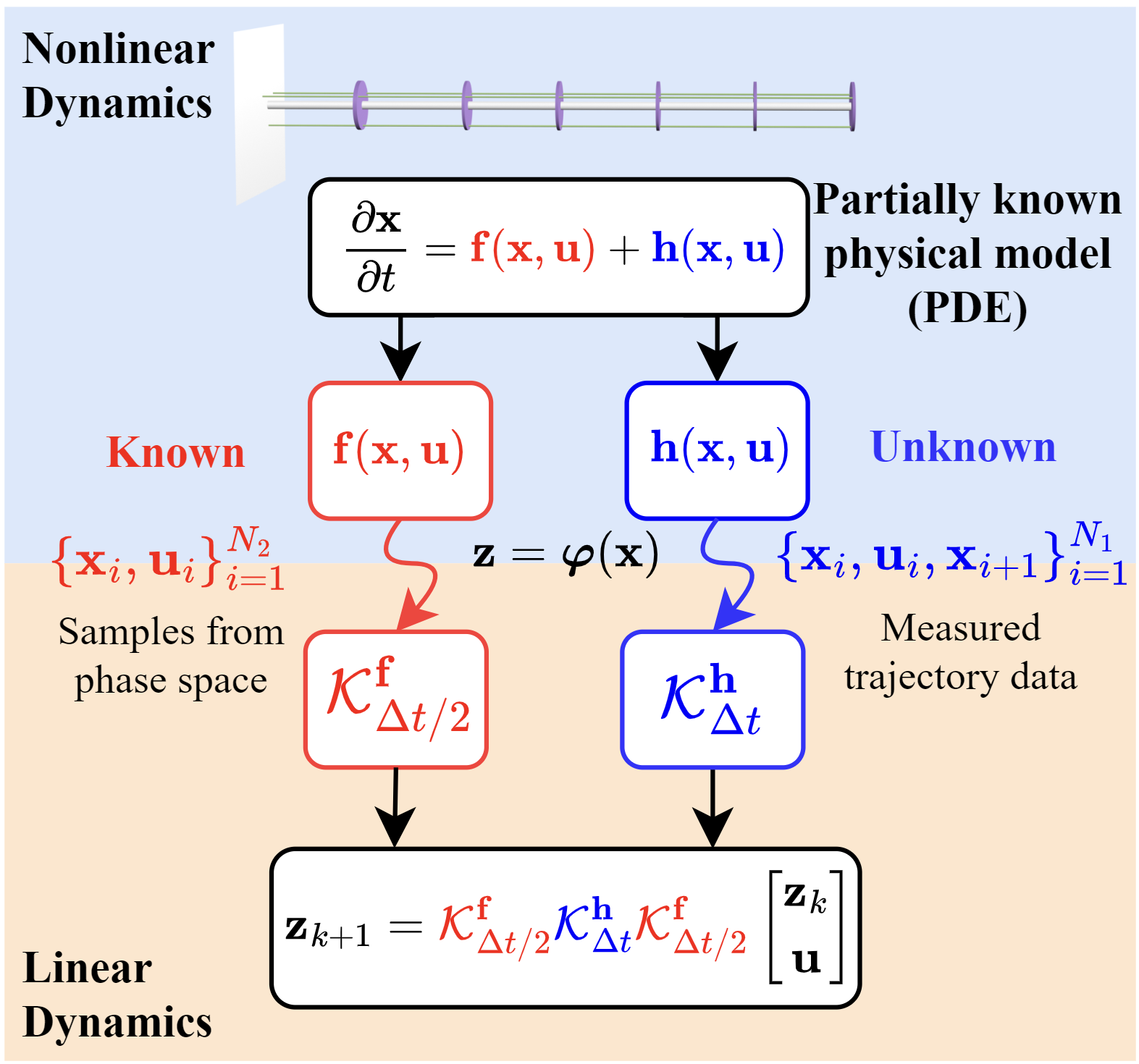}
    \caption{Flow chart depicting the PI-EDMDc approach.}
    \label{fig:flow_chart}
\vspace{-0.3in}
\end{figure}

To improve computational efficiency, data-driven methods have been proposed as a promising alternative to physics-based models. These methods minimize computational costs by using lightweight learnable operations to approximate the dynamical trajectories of a robot.
Deep neural networks have been shown to be effective \cite{thuruthel_model-based_2019, gillespie_learning_2018, zheng_robust_2020}. However, for control tasks, they cannot take advantage of explicit model-based control methods due to their black-box input-output mapping property. In contrast, Koopman operator-based methods attempt to construct globally linear models of nonlinear systems and thus can directly take advantage of well-understood linear control techniques while being computationally efficient \cite{bruder_advantages_2021, korda_linear_2018}. Especially in soft robotics, Koopman-based methods have been effectively used for soft continuum arms \cite{bruder_modeling_2019, bruder_data-driven_2021, haggerty_control_2023, shi_koopman_2023}.


Koopman operators are well-suited for modeling soft robots, but they require large amounts of data to converge. Although methods exist that identify Koopman operators in the low-data limit \cite{kaiser_sparse_2018}, in practice, tens of thousands of samples are still needed to construct an accurate model of a soft robot due to its high dimensionality, thus consuming a considerable amount of time for a physical robot to explore its configuration space~\cite{bruder_modeling_2019}.
To address this issue, physics-informed learning methods take advantage of information from known differential equations, thus reducing the high data requirement~\cite{raissi_physics-informed_2019}. For Koopman operators, neural methods that use PDEs as priors have been shown to be effective, even in the absence of true discrete data \cite{liu_physics-informed_2022}. Further, a physics-informed neural network topology can be implemented to supplement the learning of energy-conserving Koopman operators \cite{wang_physics-informed_2024}. However, unlike conventional Koopman operator identification methods, such as dynamic mode decomposition (DMD) \cite{tu_dynamic_2014} and extended dynamic mode decomposition (EDMD) \cite{williams_datadriven_2015}, neural networks are computationally expensive to train, even for low-dimensional dynamical systems.


In this work, we propose a method that decreases the required data for Koopman-based methods by incorporating physics-informed models of soft robotics arms. Specifically, as shown in Fig.~\ref{fig:flow_chart}, we present physics-informed extended dynamic mode decomposition with control (PI-EDMDc): for a system represented by PDEs, we linearly separate the PDEs into known (e.g. passive dynamics) and unknown terms (e.g. unknown actuation forces or external load/disturbance) and identify separate Koopman operators for each, enabled by an operator splitting technique known as Strang splitting. The Koopman operator associated with the known terms of the PDEs can be learned from phase space samples, while the operator associated with unknown terms must be learned from trajectory data, before being combined to produce a full linear model.  
The novel contributions of the work are as follows:
\begin{enumerate}
    \item A split operator method for partially known PDEs that is able to take advantage of phase space data, that can be obtained efficiently through solving static equations, and trajectory data, thereby significantly reducing the amount of required true trajectory data while preserving the accuracy, computational efficiency, and linearity of Koopman operator methods. 
    \item We applied the method to simulate a tendon-driven soft robotic arm where we observed nearly 3 orders of magnitude smaller shape error in comparison to existing EDMD methods when trajectory dataset sizes are small.
\end{enumerate}

The rest of this paper is organized as follows. In section \ref{sec:background}, we provide background information on Koopman operators and formulate their use for robotic systems. In section \ref{sec:methods}, we provide a description of PI-EDMDc, and practical considerations for its implementation. In section \ref{sec:sim_results}, we perform numerical experiments conducted to validate the proposed method on a soft tendon-driven continuum arm.
In section \ref{sec:conclusion}, we provide some concluding remarks.

\section{Background and Mathematical Formulation} \label{sec:background}

In this section, we provide an overview of Koopman operator theory and algorithms for data-driven identification of discrete-time and continuous-time Koopman operators. The discrete-time Koopman operator requires trajectory data while the continuous-time Koopman operator requires knowledge of the governing PDE. PI-EDMDc combines both these methods to improve approximations of Koopman operators when the trajectory dataset size is small and there exists partial knowledge of governing PDEs.

\subsection{Discrete-time Koopman Operators} \label{sec:disc_time_koopman}

\paragraph{Definitions}
Consider a dynamical system whose time evolution is described by
\begin{equation} \label{eq:x_diff_eq}
    \dot{\mathbf{x}} = \mathbf{f}(\mathbf{x}, \mathbf{u}),
\end{equation}
where $\mathbf{x} \in \mathcal{X}$ describes the state of the system evolving on some smooth manifold $\mathcal{X}$, $\dot{\mathbf{x}}$ denotes the time derivative of $\mathbf{x}$, $\mathbf{u} \in \mathcal{U}$ describes the inputs to the system on a smooth manifold $\mathcal{U}$, and $\mathbf{f}: \mathcal{X} \times \mathcal{U} \rightarrow \mathcal{X}$ is a vector field that describes the dynamics. In the context of robotics,  $\mathbf{x} \in \mathbb{R}^n$ is the state of a robot, and $\mathbf{u} \in \mathbb{R}^m$ is the applied actuation and can be assumed to not be constrained to a smooth manifold \cite{proctor_generalizing_2018}. In discrete time, the system is propagated forward in time by the flow map
\begin{align} \label{eq:x_flow_map}
\begin{split}
    \mathbf{x}(t_0 + t) &= \mathbf{F}_t(\mathbf{x}(t_0), \mathbf{u}(t_0)) \\
    &= \mathbf{x}(t_0) + \int_{t_0}^{t_0 + t} \mathbf{f}(\mathbf{x}(\tau), \mathbf{u}(\tau)) \di{\tau}.
\end{split}
\end{align}
In robotics where control inputs can be fully specified by the user, it is common to make the assumption that control inputs $\mathbf{u}$ do not evolve forward in time over a single time step \cite{bruder_modeling_2019, bruder_data-driven_2021}.

The discrete-time Koopman operator $\mathcal{K}_t$ is an infinite-dimensional linear operator that acts on functions $g: \mathcal{X} \times \mathcal{U} \rightarrow \mathbb{R}$ called \textit{observables} which belong to some infinite-dimensional Hilbert space $\mathcal{H}$. In particular, the Koopman operator propagates state measurements forward in time as
\begin{equation} \label{eq:koopman_def}
    \mathcal{K}_t g = g \circ \mathbf{F}_t,
\end{equation}
where $\circ$ is the function composition operator.

\paragraph{Discrete-time Koopman Operator Identification}
Computing the Koopman operator is infeasible on an infinite-dimensional Hilbert space. However, a Koopman-invariant subspace is spanned by a finite set of eigenfunctions of the Koopman operator, and in theory, identifying these eigenfunctions would allow us to construct a globally linear representation of the nonlinear system \cite{kaiser_data-driven_2021}. 
To identify operators associated with these eigenfunctions, we first learn approximations of the Koopman operator on a finite set of functions. Consider a dataset $D~=~\{\mathbf{x}(t_i), \mathbf{x}(t_i+\Delta t), \mathbf{u}_i\}_{i=1}^N$ and a dictionary of scalar observables $\Theta(\mathbf{x}, \mathbf{u}): \mathcal{X} \times \mathcal{U} \rightarrow \mathbb{R}^M$. Dataset $D$ yields the following data matrices: state $X$, next state $X^\prime$ after time shift $\Delta t$, and control input $U$. One can construct data matrices $\Theta(X, U)$ and $\Theta(X^\prime, U)$ where
\begin{equation} \label{eq:theta_matrix}
    \Theta(X, U) = \begin{bmatrix}
        \vertbar & & \vertbar \\
        \Theta(\mathbf{x}(t_1), \mathbf{u}_1) & \hdots & \Theta(\mathbf{x}(t_N), \mathbf{u}_N) \\
        \vertbar & & \vertbar
    \end{bmatrix},
\end{equation}
and $\Theta(X^\prime, U)$ is similarly defined, except applied to the state after time shift $\mathbf{x}(t_i + \Delta t)$. Then, the discrete-time Koopman operator can be approximated with an entirely data-driven EDMD method~\cite{williams_datadriven_2015}
\begin{align} \label{eq:discrete-time-koop}
\begin{split}
    \mathcal{K}_{\Delta t} & := \argmin_{\mathcal{K}^*_{\Delta t}} \left\|\mathcal{K}^*_{\Delta t} \Theta(X, U) - \Theta(X^\prime, U) \right\|_2^2 \\
    &\approx \Theta(X^\prime, U) \Theta^\dagger(X, U),
\end{split}
\end{align}
where $(\cdot)^\dagger$ denotes the Moore-Penrose pseudoinverse. Note that this approximation relies entirely on observed trajectory data, and does not require knowledge of the governing equations for dynamics. 

Eigenfunctions and eigenmodes can be directly extracted from $\mathcal{K}_{\Delta t}$ for explicit control over the model, but for simplicity, as done by \cite{bruder_modeling_2019, bruder_data-driven_2021}, we directly use $K_{\Delta t}$ to propagate the dictionary of functions $\Theta$ forwards in time, given by
\begin{equation} \label{eq:update_rule}
    \Theta(\mathbf{x}_{k+1}, \mathbf{u}_k) \approx \mathcal{K}_{\Delta t} \Theta(\mathbf{x}_{k}, \mathbf{u}_k).
\end{equation}

\subsection{Continuous-time Koopman Operators} \label{sec:cont_time_koopman}

\paragraph{Definitions}
The infinitesimal generator of the Koopman operator $\mathcal{L}$, known as the continuous-time Koopman operator, is also a linear operator, given by the Lie derivative
\begin{align} \label{eq:lie_deriv}
\begin{split}
    \dot{g} = \lim_{t \rightarrow 0} \frac{\mathcal{K}_t g - g}{t} = \mathcal{L} g = \nabla_\mathbf{x} g \cdot \mathbf{f} + \nabla_\mathbf{u} g \cdot \dot{\mathbf{u}},
\end{split}
\end{align}
where the last equality holds as $\mathcal{L}$ is a Liouville operator. As in the discrete-time case, we can assume that $\mathbf{u}$ is constant over a single time step, and thus its time derivative is 0.

\paragraph{Continuous-time Koopman Operator Identification}
As with the discrete-time Koopman operator, we attempt to identify operators associated with Koopman eigenfunctions that allow us to construct a globally linear model. 
Consider a dataset $D = \{\mathbf{x}_i, \mathbf{u}_i\}_{i=1}^N$ and a dictionary of scalar observables $\Theta(\mathbf{x}, \mathbf{u}): \mathcal{X} \times \mathcal{U} \rightarrow \mathbb{R}^M$. We then construct data matrix $\Theta(X, U)$ as in Eqn.~\eqref{eq:theta_matrix}. Given the PDE $\mathbf{f}(\mathbf{x}, \mathbf{u})$, one can approximate the continuous-time Koopman operator without having to rely on trajectory data using a method known as gEDMD~\cite{klus_data-driven_2020}, as 
\begin{align} \label{eq:cont_koopman_solve}
\begin{split}
    \mathcal{L} &:= \argmin_{\mathcal{L}^*} \left\|\mathcal{L}^* \Theta(X, U) - J(X, U) \right\|_2^2 \\
    &\approx J(X, U) \Theta^\dagger(X, U),
\end{split}
\end{align}
where
\begin{equation} \label{eq:jacob_solve}
    J(X, U) = \begin{bmatrix}
        \vertbar & & \vertbar \\
        J(\mathbf{x}(t_1), \mathbf{u}_1) & \hdots & J(\mathbf{x}(t_N), \mathbf{u}_N) \\
        \vertbar & & \vertbar 
    \end{bmatrix},
\end{equation}
and
\begin{equation}
    J(\mathbf{x}, \mathbf{u}) = J_{\mathbf{x}} \Theta(\mathbf{x}, \mathbf{u}) \cdot \mathbf{f}(\mathbf{x}, \mathbf{u}),
\end{equation}
where $J_{\mathbf{x}}\Theta$ is the Jacobian matrix of $\Theta$ with respect to $\mathbf{x}$, defined by
\begin{equation}    J_{\mathbf{x}}\Theta(\mathbf{x}, \mathbf{u}) = \begin{bmatrix}
        \vertbar & & \vertbar \\
        \nabla_{\mathbf{x}} \Theta_1 (\mathbf{x}, \mathbf{u}) & \hdots & \nabla_{\mathbf{x}} \Theta_M(\mathbf{x}, \mathbf{u}) \\
        \vertbar & & \vertbar
    \end{bmatrix}^\intercal.
\end{equation}
The approximation of the continuous-time Koopman operator relies entirely on governing PDEs, and as such, the dataset does not need to be sampled from trajectories, but instead, points can be sampled freely from phase space using techniques such as Latin hypercube sampling \cite{kaiser_data-driven_2021}. This fact is a strong motivator for PI-EDMDc, as given terms of PDEs, we are able to learn associated Koopman operators without relying on trajectories, thereby reducing the amount of data needed to resolve nonlinearities arising due to those terms. 

Unlike in the discrete-time case, the continuous time Koopman operator cannot be used directly to propagate the dictionary of functions forward in time. Instead, its corresponding solution operator can be computed from any approximations of the matrix exponential \cite{moler_nineteen_2003} given by
\begin{equation} \label{eq:matrix_exp}
    \mathcal{K}_{\Delta t} = e^{\Delta t \mathcal{L}}.
\end{equation}
In particular, we use the scaling and squaring method developed by \cite{al-mohy_new_2009}.

\subsection{Control-affine and Bilinear Koopman Operator Identification}

For generality, the Koopman operator identification techniques described in Sec.~\ref{sec:disc_time_koopman} and \ref{sec:cont_time_koopman} solve for the Koopman operator by applying a linear operator to an arbitrary and potentially nonlinear dictionary $\Theta(\mathbf{x}, \mathbf{u})$. More principled choices can be adopted 
that align better with standard control algorithms such as model predictive control (MPC). 

Following \cite{proctor_generalizing_2018}, we can split the observables into two separate Hilbert spaces, $\mathcal{H}_{\mathbf{x}}$ and $\mathcal{H}_{\mathbf{xu}}$, where $g_{\mathbf{x}} \in \mathcal{H}_{\mathbf{x}}: \mathcal{X} \rightarrow \mathbb{R}$ and $g_{\mathbf{xu}} \in \mathcal{H}_{\mathbf{xu}}: \mathcal{X} \times \mathcal{U} \rightarrow \mathbb{R}$. The full Hilbert space is considered, without loss of generality, as their composition, $\mathcal{H} = \mathcal{H}_{\mathbf{x}} \otimes \mathcal{H}_{\mathbf{xu}}$. Under this separation, a \textit{vector} of scalar observables $\mathbf{g} = \begin{bmatrix}
    \mathbf{g}_{\mathbf{x}}^\intercal & \mathbf{g}_{\mathbf{xu}}^\intercal
\end{bmatrix}^\intercal$, where $g^i_\mathbf{x} \in \mathcal{H}_\mathbf{x}$ and $g^i_\mathbf{xu} \in \mathcal{H}_\mathbf{xu}$, has a time evolution described by
\begin{equation} \label{eq:hilbert_split}
    \begin{bmatrix}
        \dot{\mathbf{g}}_{\mathbf{x}} \\ \dot{\mathbf{g}}_{\mathbf{xu}}
    \end{bmatrix} = 
    \begin{bmatrix}
        \mathcal{L}_{11} & \mathcal{L}_{12} \\
        \mathcal{L}_{21} & \mathcal{L}_{22}
    \end{bmatrix} 
    \begin{bmatrix}
        \mathbf{g}_{\mathbf{x}} \\ \mathbf{g}_{\mathbf{xu}}
    \end{bmatrix},
\end{equation}
which, under integration over a small period of time, has a similar form for the discrete-time analogue.

Linear extended dynamic mode decomposition with control (L-EDMDc) \cite{korda_linear_2018} identifies Koopman operators with a dictionary satisfying the form
\begin{equation} \label{eq:l_edmdc_dict}
    \Theta(\mathbf{x}, \mathbf{u}) = \begin{bmatrix}
        \Theta_{\mathbf{x}}(\mathbf{x}) \\
        \mathbf{u}
    \end{bmatrix},
\end{equation}
where $\Theta_{\mathbf{x}} : \mathcal{X} \rightarrow \mathbb{R}^{M-m}$ is a dictionary of scalar observables in $\mathcal{H}_\mathbf{x}$ and we note that $\mathbf{u} \in \mathbb{R}^m$. This method identifies models that explicitly preserve linearity of control inputs, making it straightforward to apply techniques such as LQR or linear MPC.

Bilinear extended dynamic mode decomposition with control (B-EDMDc)~\cite{bruder_advantages_2021} argues in favor of a bilinear Koopman realization that learns unknown dynamics by using a dictionary that satisfies the form
\begin{equation}
    \Theta(\mathbf{x}, \mathbf{u}) = \begin{bmatrix}
        \Theta_{\mathbf{x}}(\mathbf{x}) \\
        u_1 \Theta_{\mathbf{x}}(\mathbf{x}) \\
        \vdots \\
        u_m \Theta_{\mathbf{x}}(\mathbf{x}) \\
    \end{bmatrix}.
\end{equation}
This method preserves a bilinear structure with respect to control inputs, making it applicable to efficient optimal control for bilinear dynamical systems \cite{aganovic_linear_1995, bruder_advantages_2021}.

\section{Methodology} \label{sec:methods}

Both L-EDMDc and B-EDMDc rely on the discrete-time Koopman operator approximation, which requires trajectory data to resolve. As shown in Sec.~\ref{sec:cont_time_koopman}, the continuous-time Koopman operator can be learned directly from phase space. Motivated by this, we investigate how one might combine continuous-time and discrete-time Koopman operators. In this section, we introduce the theory of and a practical algorithm for PI-EDMDc, which learns Koopman operators for partially known and linearly separable PDEs.

\subsection{Koopman Operators for Partially Known PDEs}
A common choice for modeling continuum soft robots is Cosserat-rod theory~\cite{rucker_statics_2011, till_real-time_2019, renda_dynamic_2014}, whose governing equations can be linearly separated into a passive system and actuation forces, which will be introduced in Sec.~\ref{sec:tendon_robot}. In such cases, the passive (known) system is well understood, whereas the actuated system may have non-trivial coupling with state stemming from unknown forces. Consider PDEs such as this, in the form
\begin{equation} \label{eq:linearly_separable_diffeq}
    \dot{\mathbf{x}} = \mathbf{f}(\mathbf{x}, \mathbf{u}) + \mathbf{h}(\mathbf{x}, \mathbf{u}),
\end{equation}
where $\mathbf{f}: \mathcal{X} \times \mathcal{U} \rightarrow \mathcal{X}$ is known and $\mathbf{h}: \mathcal{X} \times \mathcal{U} \rightarrow \mathcal{X}$ is unknown. In the case of robots with dynamics partially described by Cosserat-rod theory, the passive backbone dynamics do not depend on control inputs, i.e. $\mathbf{f}(\mathbf{x}, \mathbf{u}) \equiv \mathbf{f}(\mathbf{x})$. 

To approximate the full Koopman operator of the linearly separable differential equation as a composition of a continuous-time and a discrete-time Koopman operator, we make use of a technique called \textit{Strang splitting}, a common operator splitting technique that splits the Liouvillian and results in a second-order error. Specifically, consider the linearly separable Liouvillian of Eqn.~\eqref{eq:linearly_separable_diffeq}, given by:
\begin{equation}
    \mathcal{L}^{\mathbf{f}+\mathbf{h}} = \mathcal{L}^\mathbf{f} + \mathcal{L}^\mathbf{h},
\end{equation}
where the superscript denotes the function corresponding to the Liouville operator. The full solution operator can be derived from the Baker-Campbell-Hausdorff (BCH) expansion:
\begin{equation}
	e^{\mathcal{L}^{\mathbf{f}+\mathbf{h}} t} = e^{\mathcal{L}^\mathbf{f} t} e^{\mathcal{L}^\mathbf{h} t} + \mathcal{O}(t^2).
\end{equation}
Strang splittings involve a symmetric splitting, which, as a numerical integration scheme, results in second order accuracy with respect to the size of the time step $\Delta t$
\begin{equation}
	e^{\mathcal{L}^{\mathbf{f}+\mathbf{h}} \Delta t} \approx e^{\mathcal{L}^\mathbf{f} \Delta t/2} e^{\mathcal{L}^\mathbf{h} \Delta t} e^{\mathcal{L}^\mathbf{f} \Delta t/2}, 
\end{equation}
where the updates due to $\mathcal{L}^f$ and $\mathcal{L}^h$ can be interchanged.

The continuous time evolution of an observable satisfies
\begin{equation}
    \dot{g} = \mathcal{L} g = \nabla_{\mathbf{x}} g \cdot \mathbf{f} + \nabla_{\mathbf{x}} g \cdot \mathbf{h}.
\end{equation}
As such, when applying a coordinate transform to the Koopman eigenbasis of the full system, 
the full Koopman operator can be approximated using Strang splitting as
\begin{equation} \label{eq:koopman_strang_splitting}
	\mathcal{K}_t \approx \mathcal{K}_{t/2}^\mathbf{f} \mathcal{K}_{t}^\mathbf{h} \mathcal{K}_{t/2}^\mathbf{f},
\end{equation}
where the superscript denotes the term for which the Koopman operator provides a solution operator. 

\begin{algorithm}[h]
\caption{PI-EDMDc}\label{alg:pi-edmdc}
    \SetKwInOut{Input}{Input}

    
    \Input{Chosen dictionary of functions $\Theta(\mathbf{x}, \mathbf{u})$}
    
    \KwData{Trajectories: $D_1 = \{ \mathbf{x}(t_i), \mathbf{x}(t_i + \Delta t), \mathbf{u}(t_i) \}_{i=1}^{N_1}$. \qquad~~Phase space samples: $D_2 = \{ \mathbf{x}_i, \mathbf{u}_i \}_{i=1}^{N_2}$}
    \KwResult{$\Theta(\mathbf{x}_{k+1}, \mathbf{u}_k) \approx \mathcal{K}_t \Theta(\mathbf{x}_k, \mathbf{u}_k)$}
    
    $\mathcal{L}^\mathbf{f} \approx J(X_2, U_2) \Theta^\dagger(X_2, U_2)$ \Comment*[r]{Eqn.~\eqref{eq:cont_koopman_solve}}
    

    $\mathcal{K}_{t/2}^\mathbf{f} \gets e^{t \mathcal{L}^{\mathbf{f}} /2}$ \Comment*[r]{Eqn.~\eqref{eq:matrix_exp}}

    $\mathcal{K}_{t}^\mathbf{h} \gets (\mathcal{K}_{t/2}^\mathbf{f})^\dagger \Theta(X^\prime_1, U_1) (\mathcal{K}_{t/2}^\mathbf{f} \Theta(X_1, U_1))^\dagger$ \Comment*[r]{Eqn.~\eqref{eq:full_kth}}

    $\mathcal{K}_t \gets \mathcal{K}_{t/2}^\mathbf{f} \mathcal{K}_{t}^\mathbf{h} \mathcal{K}_{t/2}^\mathbf{f}$ \Comment*[r]{Eqn.~\eqref{eq:koopman_strang_splitting}}
\end{algorithm}


\subsection{Physics Informed EDMDc}

Consider a dataset of trajectories, $D_1 = \{ \mathbf{x}(t_i), \mathbf{x}(t_i + \Delta t), \mathbf{u}(t_i) \}_{i=1}^{N_1}$, and a dataset corresponding to collocation points on the manifold $\mathcal{X} \times \mathcal{U}$, $D_2 = \{ \mathbf{x}_i, \mathbf{u}_i \}_{i=1}^{N_2}$. In practice, especially in online settings, the size of $D_1$ may increase slowly as true trajectories must be collected in real time, while $D_2$ can be easily obtained as in Sec \ref{sec:tendon_robot} (and therefore can be arbitrarily large). It is a natural choice then to learn the known part $\mathcal{K}_{t/2}^\mathbf{f}$ entirely from phase space samples, and the unknown part $\mathcal{K}_{t}^\mathbf{h}$ from the trajectory data.

As summarized in Alg.~\ref{alg:pi-edmdc}, given a dictionary of scalar observables $\Theta(\mathbf{x}, \mathbf{u}): \mathcal{X} \times \mathcal{U} \rightarrow \mathbb{R}^M$, we can learn the continuous-time Koopman operator associated with $\mathbf{f}(\mathbf{x}, \mathbf{u})$ by following Eqn.~\eqref{eq:cont_koopman_solve} and using dataset $D_2$. Using Eqn.~\eqref{eq:matrix_exp}, we can construct a corresponding discrete-time Koopman operator $\mathcal{K}_{t/2}^f$ over the interval $\Delta t / 2$. 

As in Eqn.~\eqref{eq:theta_matrix}, we can construct data matrices $\Theta(X^\prime, U)$ and $\Theta(X, U)$ from the trajectory data in dataset $D_1$, and with $\mathcal{K}_{t/2}^\mathbf{f}$ derived from the previous step, we compute
\begin{align} \label{eq:full_kth}
\begin{split}
    \begin{split}
    \mathcal{K}_{t}^{\mathbf{h}} &:= \argmin_{\mathcal{K}_{t}^{\mathbf{h}*}} \|  \mathcal{K}_{t}^{\mathbf{h}*} \mathcal{K}_{t/2}^\mathbf{f} \Theta(X, U) - \\
    &\phantom{:= \argmin\|~}(\mathcal{K}_{t/2}^\mathbf{f})^\dagger \Theta(X^\prime, U) \|_2^2 
    \end{split}\\
    &\approx (\mathcal{K}_{t/2}^\mathbf{f})^\dagger \Theta(X^\prime, U) (\mathcal{K}_{t/2}^\mathbf{f} \Theta(X, U))^\dagger.
\end{split}
\end{align}
The minimization problem in Eqn.~\eqref{eq:full_kth} can be derived by rearranging the equality
\begin{equation}
    \mathcal{K}_{t/2}^\mathbf{f} \mathcal{K}_{t}^\mathbf{h} \mathcal{K}_{t/2}^\mathbf{f} \Theta(X, U) = \Theta(X^\prime, U),
\end{equation}
which can also be used to approximately propagate the system of measurements forward in time. A summary of the full procedure is described by Alg.~\ref{alg:pi-edmdc}.

\subsection{Practical Considerations} \label{sec:prac_considerations}

\paragraph{Evaluation of $\mathcal{K}_{t/2}^\mathbf{f}$}

If the eigenvalues of the continuous-time Koopman operator identified by step 1 of Alg.~\ref{alg:pi-edmdc} are spurious, then the matrix exponentiation in step 2 will result in an unstable matrix. The discrete-time Koopman operator can instead be learned directly from the forward flow map associated with $\mathbf{f}$:
\begin{equation}
    \mathbf{F}_{t}^\mathbf{f}(\mathbf{x}(t_0), \mathbf{u}) = \mathbf{x}(t_0) + \int_0^t \mathbf{f}(\mathbf{x}(\tau), \mathbf{u}) \di{\tau},
\end{equation}
and using the solution given by Eqn.~\eqref{eq:discrete-time-koop}. If computing $\mathbf{F}_{t}^f$ exactly is not numerically feasible, first-order approximations of its solution may suffice.

\paragraph{Regularization}

The pseudoinverse in Eqn.~\eqref{eq:discrete-time-koop}, \eqref{eq:cont_koopman_solve}, and~\eqref{eq:full_kth} is prone to overfitting in low data limits, and is highly sensitive to outliers and noisy data during training \cite{seheult_robust_1989}. To mitigate this issue,  we regularize the matrices obtained while minimizing the $L^2$ error using a method suggested by Bruder et. al. \cite{bruder_modeling_2019}. Specifically, we apply an $L^1$ regularization using the Least Absolute Shrinkage and Selection Operator (LASSO) \cite{tibshirani_regression_1996}. This approach improves the sparsity of learned matrices as the $L^1$ regularization is capable of driving terms to 0, and also tends to reduce the magnitude of eigenvalues, thereby improving the stability of the learned system. 
For each least squares minimization in the form $AX = X^\prime$, we add an $L^1$ regularization as
\begin{equation}
    A := \argmin_{A} \left(\|A X - X^\prime \|_2^2 + \alpha \|A\|_1\right),
\end{equation}
where $\alpha$ is a hyperparameter that controls the magnitude of the regularization term.

\paragraph{Adjusting the full $\mathcal{K}_t$}

In the case where certain terms of $\mathbf{h}$ are known to be small or zero, such as in Eqn.~\eqref{eq:cosserat_rod_eq}, it may be desirable to directly use the corresponding solutions approximated merely by $\mathcal{K}_{t/2}^\mathbf{f}$ applied twice, which may in the full formulation be obfuscated by the numerical procedures required in computing step 4 of Alg.~\ref{alg:pi-edmdc}. This obfuscation may be further exacerbated by the scaling of dynamical variables after the first application of $\mathcal{K}_{t/2}^\mathbf{f}$, and by the tendency of Eqn.~\eqref{eq:full_kth} to account for the second order errors in the operator splitting which may be noisy when $|D_1|$ is small. To address this, the rows of $\mathcal{K}_t^\mathbf{h}$ which correspond to the small or zero values in $\mathbf{h}$ may be re-weighted or ignored in the optimization problem described by Eqn.~\eqref{eq:full_kth}.

\section{Simulation Results} \label{sec:sim_results}

In this section, we perform numerical experiments involving a simulated soft robotic arm, and validate the effectiveness of PI-EDMDc over L-EDMDc and B-EDMDc over the simulated data. 


\subsection{Tendon-driven Soft Robot Setup} \label{sec:tendon_robot}

\begin{figure}
    \centering
    \includegraphics[width=1\linewidth]{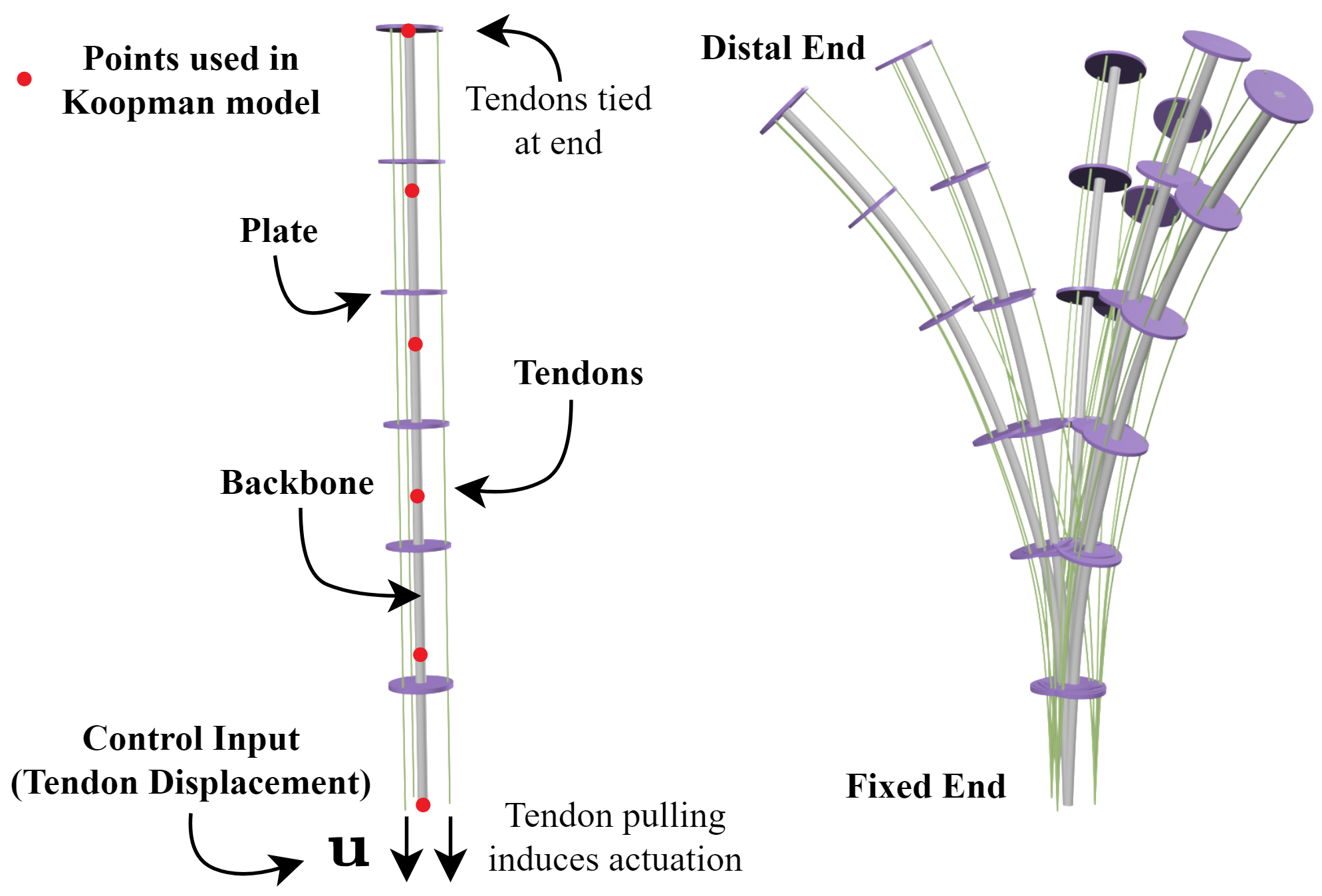}
    \caption{On the left, the reference configuration of the tendon robot setup is shown in the absence of gravity. The right showcases various configurations of the tendon robot after two of its tendons have been actuated.}
    \label{fig:tendon-configurations}
\vspace{-0.15in} 
\end{figure}

\begin{figure*}
    \normalsize

    \begin{equation} \label{eq:cosserat_rod_eq}
    \begin{bmatrix}
        \dot{\mathbf{p}} \\
        \dot{R} \\
        \dot{\mathbf{v}} \\
        \dot{\mathbf{w}} \\
        \dot{\mathbf{q}} \\
        \dot{\boldsymbol{\omega}}
    \end{bmatrix} = \underbrace{\begin{bmatrix}
        R \mathbf{q} \\
        R \hat{\boldsymbol{\omega}} \\
        \mathbf{q}_s + \hat{\mathbf{w}} \mathbf{q} - \hat{\boldsymbol{\omega}} \mathbf{v} \\
        \boldsymbol{\omega}_s + \hat{\mathbf{w}} \boldsymbol{\omega} \\
        \frac{1}{\rho A} \left[
            K_{\text{se}} (\mathbf{v}_s - \mathbf{v}_s^*) + \hat{\mathbf{w}} K_{\text{se}} (\mathbf{v} - \mathbf{v}^*) - \rho A \hat{\boldsymbol{\omega}} \mathbf{q}
        \right] \\
        (\rho J)^{-1} \left[
            K_{\text{bt}} (\mathbf{w}_s - \mathbf{w}_s^*) + \hat{\mathbf{w}} K_{\text{bt}} (\mathbf{w} - \mathbf{w}^*) + \hat{\mathbf{v}} K_{\text{se}} (\mathbf{v} - \mathbf{v}^*) - \hat{\boldsymbol{\omega}} \rho J \boldsymbol{\omega}
        \right]
    \end{bmatrix}}_{\mathbf{f}~\text{(known)}} + \underbrace{\begin{bmatrix}
        \mathbf{0} \\ \mathbf{0} \\ \mathbf{0} \\ \mathbf{0} \\ \frac{1}{\rho A} R^\intercal (\mathbf{f}_e + \mathbf{f}_t) \\ (\rho J)^{-1} R^\intercal (\mathbf{l}_e + \mathbf{l}_t)
    \end{bmatrix}}_{\mathbf{h}~\text{(unknown)}}
    \end{equation}

    \hrulefill
\end{figure*}

\begin{figure*}
    \centering
    \includegraphics[width=1\linewidth]{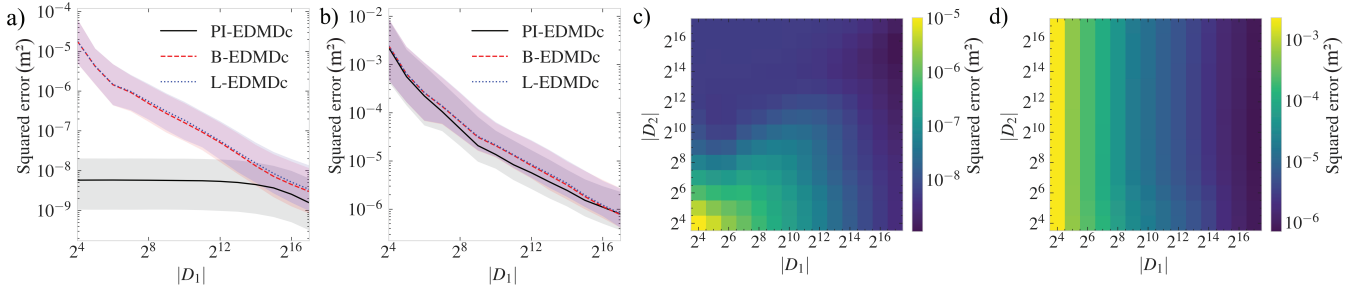}
    \caption{(a,b) Median and 25th through 75th percentile range of the squared error of various EDMDc models over the entire shape of the tendon robot (a) and the velocity at the distal end (b) as a function of $|D_1|$, the size of the collected trajectory data. (c,d) The median squared error as a function of training set size $|D_1|$ and phase space samples $|D_2|$ of the entire shape of the tendon robot (c) and the velocity at the distal end (d).} 
    \vspace{-15pt} 
    \label{fig:tendon_results_a}
\end{figure*}

The robot we use for our numerical experiments is a tendon-driven robot with a continuum backbone (Fig.~\ref{fig:tendon-configurations}). For simplicity, we choose its initial configuration to be straight. There are three tendons routed around the backbone, at 0, 120, and 240 degrees. One end of the tendon robot is fixed, while the distal end can move freely. Cosserat rod material parameters and geometry in \cite{till_real-time_2019} are used.

The dynamical equations used in the simulations are based on Cosserat rod theory, and can be described with PDEs Eqn.~\eqref{eq:cosserat_rod_eq} \cite{rucker_statics_2011}. Here, $\mathbf{p}$ and $R$ describe the position and orientation of each cross section along the rod, $\mathbf{v}$ and $\mathbf{w}$ are longitudinal and torsional strain respectively, and $\mathbf{q}$ and $\boldsymbol{\omega}$ are linear and angular velocities. $\mathbf{f}_e$, $\mathbf{l}_e$, $\mathbf{f}_t$ and $\mathbf{l}_t$ correspond to external forces and moments and tendon forces and moments. In the simulation, we use displacements of each tendon as input ($\mathbf{u} \in \mathbb{R}^3$), and therefore, tendon forces are implicit functions of $\mathbf{u}$. These dynamical variables represent an infinitesimal segment along the rod, and as such, are functions of arclength $s$. The subscript $(\cdot)_s$ denotes the partial derivative with respect to arclength. For a description of the remaining variables and a discussion of the derivation of the above equation, we refer the reader to~\cite{rucker_statics_2011}. 

To satisfy the requirements made by PI-EDMDc in Eqn.~\eqref{eq:linearly_separable_diffeq}, we linearly separate the governing Eqn.~\eqref{eq:cosserat_rod_eq} into the known (passive backbone dynamics), and unknown terms (the external forces and moments due to tendon actuation). In general, and especially in soft robotics, it is difficult to characterize all unknown external forces due to difficulties arising from the capabilities of sensor measurements. In such cases, splitting the governing equations in this way has implications for online force identification, a consequence which will be investigated further in future work.


The soft robot PDEs are numerically solved using the method proposed by \cite{till_real-time_2019} with a timestep of $0.03$ seconds, and are discretized to 101 points. We also solve the corresponding static equilibrium ODEs with the same discretization using a shooting method as done in \cite{till_real-time_2019}.
The trajectory dataset $D_1$ is collected from numerical solutions of the governing PDEs, while the phase space dataset $D_2$ is collected by solving the ODEs. The linear and angular velocities required for the phase space samples cannot be obtained from the static equilibrium solutions, and as such, these values are sampled from a multivariate Gaussian distribution with zero mean and variance determined by the full trajectory simulations. 

When the method is used with a physical prototype, $D1$ can be directly collected from experimental results, while $D_2$ can still be collected from ODE solutions. Experimental results \cite{rucker_statics_2011} suggest that, for well characterized physical prototypes, numerical solutions to the static equilibrium equations have high prediction accuracy. Further, since the static equilibrium equations are ODEs, their solutions can be efficiently numerically computed. As such, we can practically compute arbitrarily large datasets $D_2$ that sample a representative subset of the full phase space of the robot. In high inertial regimes, the static equilibrium solutions represent a smaller subset of the full manifold, and more principled ways of sampling phase space may be required.


To reduce the dimension of the discretized state, 6 of the 101 discretized points in the simulation were taken uniformly along the arclength of the tendon robot to construct the state vector $\mathbf{x}$. Each point is represented by a 24-dimensional vector of the dynamical variables, resulting in the full discretized state $\mathbf{x} \in \mathbb{R}^{144}$. 700 simulations of 200 time steps each were conducted for $D_1$, and an additional 140,000 configurations of the tendon robot were obtained for $D_2$. A LASSO regularization was used with an $\alpha$ value of $0.01 N$ where $N$ is the size of dataset used in the corresponding minimization problem. Similar to \cite{haggerty_control_2023}, we choose $\Theta(\mathbf{x}, \mathbf{u})$ to be 4 time-delayed measurements \cite{brunton_chaos_2017} with respect to $\mathbf{x}$ and the identity function over $\mathbf{u}$, in a form similar to Eqn.~\eqref{eq:l_edmdc_dict}. The dictionary of functions for linear and bilinear EDMDc are also chosen to be 4 time-delayed measurements. Consequently, we have $M = 723$, and $\mathcal{K}_{t/2}^\mathbf{f}, \mathcal{K}_{t}^\mathbf{h} \in \mathbb{R}^{723 \times 723}$, made square by appending a $579 \times 579$ identity matrix and a $579 \times 144$ matrix of zeros conformable with the time-delay coordinates and control inputs.

\subsection{Results and Discussion}
To evaluate each method, an additional 50 simulations of 200 timesteps each were conducted. We use the shape error of the robot and the velocity at the distal end to quantify the effectiveness of the method. Shape error is computed as the squared error of the concatenated vector of discretized positions along the rod $\|\mathbf{x}_{\mathbf{p},i}^{\text{true}} - \mathbf{x}_{\mathbf{p},i}^{\text{pred}}\|^2$. The distal velocity error is computed as the squared error of the velocity at the distal end of the robot. Fig.~\ref{fig:tendon_results_a}a-b show the median shape error along the rod and the velocity error of the distal end as a function of the training set size $|D_1|$. Notably, the shape error of PI-EDMDc is nearly 3 orders of magnitude smaller than B-EDMDc or L-EDMDc when $|D_1|$ is small, but does not significantly improve until $|D_1| > 2^{12}$. This gradual improvement of the method may in part be due to the large amount of data necessary to accurately recover the second order errors resulting from the splitting of the solution operators. 
Fig.~\ref{fig:tendon_results_a}c-d show the median squared error of the position along the rod and the error of the velocity at the distal end as a function of $|D_1|$ and $|D_2|$. We observe that the estimation of velocity is largely unaffected by the size of the set of phase space samples $|D_2|$. This implies that the information gained by knowledge of the passive velocity dynamics is significantly smaller than the information necessary to resolve the nonlinear forces applied due to tendon actuation and external forces. This result is not surprising, as these forces largely determine the time evolution of linear and angular velocity. The relative magnitudes of information contributed by $\mathbf{f}$ and $\mathbf{h}$ from Eqn.~\eqref{eq:linearly_separable_diffeq} on observed trajectories greatly impacts the effectiveness of the method in estimating certain dynamical variables. A rigorous characterization of this property is left to future work.




\section{Conclusion} \label{sec:conclusion}

In this paper, we have proposed a new Koopman operator identification method that takes advantage of partial knowledge of governing PDEs, enabling the use of phase space samples facilitated by a Strang splitting to enhance learning of Koopman operators for the forward estimation of robotic systems. We show that, on numerically simulated data, this method is able to improve simulation accuracy of certain dynamical variables by orders of magnitude in small data limits, while preserving the linear structure of the resulting model. In future work, we would like to apply this technique to data collected from physical prototypes of soft robotic arms and apply linear control techniques such as model predictive control to learned models to identify the benefits of using PI-EDMDc in practical settings.





\bibliographystyle{IEEEtran}
\bibliography{bst,refs}
%

\end{document}